\documentclass[journal]{IEEEtran}

\usepackage[utf8]{inputenc}
\usepackage[T1]{fontenc}
\usepackage{microtype}
\usepackage{booktabs}
\usepackage{array}
\usepackage{xcolor}
\usepackage{amsmath}
\usepackage{graphicx}
\definecolor{okblue}{RGB}{0,114,178}
\definecolor{okorange}{RGB}{230,159,0}
\definecolor{okverm}{RGB}{213,94,0}
\definecolor{okgreen}{RGB}{0,158,115}
\usepackage[hidelinks]{hyperref}
\usepackage{xurl}
\usepackage{needspace}

\newcommand{\code}[1]{\texttt{#1}}
\newcommand{\fire}{$\bullet$}
\newcommand{\silent}{$\circ$}
\title{Deterministic LLM Inference Across GPU Kernels:\\
Power-of-Two INT8 Quantization Scales and\\
the Limits of Tolerance-Based Conformance}

\author{Teng-Ruei~Chen%
\thanks{T.-R. Chen is with Krixvon, Taipei 100, Taiwan (e-mail: luka@krixvon.com; ORCID: 0000-0001-8995-334X).}%
\thanks{Preprint, August 2026. Companion to arXiv:2608.13756, whose open questions on check sensitivity and on the deployability of power-of-two scales this paper answers; text overlap with the companion is negligible and every result here is new. All measurements ran on a pinned software stack on a single RTX 4090 under a pre-registered protocol with append-only amendments.}}

\begin{document}
\maketitle

\begin{abstract}
INT8 quantization is now a standard serving configuration for large language models, and
two GPU kernels that implement the same scaled INT8 GEMM interface are ordinarily treated
as interchangeable. A companion study (arXiv:2608.13756) showed they are not
interchangeable bitwise, localized their disagreement to the epilogue---the scale
multiplications and final rounding after an exact integer accumulator---and distilled its
controls into a conformance suite; however, that suite was only ever run on presumed-good
kernels, so its checks passed without ever being shown to fire, and the same work reported
$+157\%$ perplexity for its power-of-two scale probe, leaving the one bitwise-strict check
without a servable checkpoint to run on. This paper measures both gaps. Measuring check
sensitivity requires faults whose ground truth is known, so we build a reference pipeline
whose accumulator is exact by construction and inject nine fault families---five plausible
epilogue defects, two precondition violations, an operand mismatch, and a null---across
196 captured layers of Qwen3-1.7B, three coverage severities, and two scale regimes:
8{,}232 cells, each scored against a 77-cell prediction matrix whose 63-cell core was
fixed before any data; a disclosed two-layer smoke run expanded it to 77 cells and changed
three predictions, and the corrected matrix was re-pinned before the full run. The suite
records zero false positives and zero false negatives against the corrected matrix, yet four of the five epilogue faults are detected by no check in any cell:
every epilogue fault moves an output by at most one bfloat16 spacing---exactly one
whenever it moves it at all---so a tolerance of one spacing is blind to the class by
construction, while an operand mismatch reaches distances in the tens of thousands.
Deployability fails for a different reason than reported: requantizing from the parent
weights under power-of-two scales---validated by rebuilding the committed checkpoint byte
for byte under the unconstrained rule---attributes $99.8\%$ of the probe's $+157\%$ to a
weight--scale mismatch in the probe's construction rather than to the constraint. The
requantized checkpoints make CUTLASS and Triton agree bitwise at every linear layer
(196/196 and 252/252, against 8/196 and 10/252 under the checkpoints' own scales) and
yield byte-identical greedy token sequences at 1.7B, 8B, and 14B (8/8 prompts at all
three, against 0/8), at perplexity point estimates of $+0.32\%$, $-0.28\%$, and $+0.48\%$ (the $90\%$
interval covers zero at the two smaller sizes but not at 14B, and its upper ends reach
$+0.71\%$ and $+0.76\%$). Power-of-two scales are thus a deployable serving configuration for cross-kernel
bitwise determinism, and the condition under which this suite's epilogue comparison
becomes a required equality rather than a tolerance: what a tolerance-based suite of this
shape honestly certifies is preconditions, operand provenance, and one-spacing
boundedness---not interchangeability.
\end{abstract}


\begin{figure*}[!t]
\centering
\includegraphics[width=\textwidth]{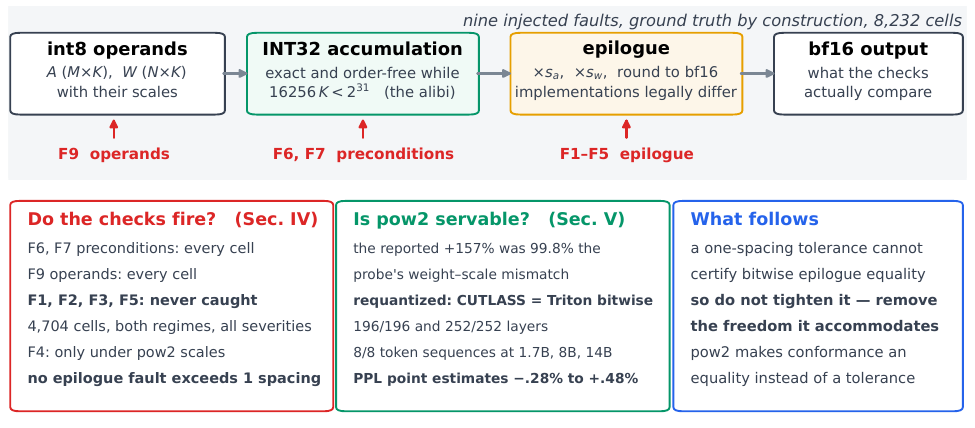}
\caption{Overview. \emph{Top:} a W8A8 linear layer's integer accumulation is exact and
order-independent under a verified bound, so cross-kernel differences can arise only in the
epilogue; nine fault families are injected with ground truth known by construction and
scored against a prediction matrix (63 cells pre-data; corrected and re-pinned after a
disclosed smoke run). \emph{Bottom left:} the suite is
reliable about its own preconditions and about operand provenance, and structurally blind to
epilogue defects that stay within one bfloat16 spacing---which is where single-rounding and
precision faults live. \emph{Bottom centre:} power-of-two scales, built by requantization
rather than scale rewriting, give bitwise cross-kernel agreement at observed perplexity costs between $-0.28\%$ and
$+0.48\%$.}
\label{fig:overview}
\end{figure*}

\section{Introduction}

Two GPU kernels that implement the same scaled INT8 GEMM interface are
ordinarily treated as interchangeable, and production serving stacks treat the choice
between them as an implementation detail: vLLM selects between a CUTLASS and a Triton
path for the same quantized linear layer by hardware and heuristics~\cite{vllm,cutlass,
tillet2019triton}, and INT8 W8A8 remains a standard configuration for serving large
language models~\cite{jacob2018integer,wu2020integer,qwen3}. Whether that choice is
\emph{observable} in the output is a reproducibility question with practical stakes:
regression tests, cached evaluations, audit trails, and any pipeline that compares logits
across deployments all assume the kernel choice washes out.

It does not, and this paper is a direct continuation of the study that established
that---\emph{The Integer Alibi}~\cite{integeralibi2026}, \code{arXiv:2608.13756}---so it
is worth stating at the outset what that study found and what it left open. Holding the
checkpoint, prompts, hardware, engine, decoding, and quantization configuration fixed and
swapping only the INT8 linear kernel, it found each arm bit-reproducible against itself
while the two arms agreed on \emph{no} generated sequence. It then localized the
disagreement: because an INT8$\times$INT8 product accumulated in INT32 is exact and
order-independent under a no-overflow bound, the accumulator cannot cause any cross-kernel
difference---an \emph{integer alibi}---so whatever differs must differ in the epilogue,
where scales are applied and the result rounds to bfloat16
(Fig.~\ref{fig:overview}, top). From those controls it distilled a seven-check conformance
suite, and used a power-of-two scale rewrite as a diagnostic probe that restored bitwise
agreement end to end. This sits within a line of work on nondeterminism in LLM
inference~\cite{schlogl2023deviations,yuan2025nondeterminism,he2025batchinvariance,
tbik2025} and an older tradition of testing numerical
kernels~\cite{pham2019cradle,zhang2021predoo,dubey2025volta,valpey2025smt}, reviewed in
Section~\ref{sec:related}.

However, that study left its two operational conclusions unproven, and said so. First, the
suite ran only on presumed-good kernels: ``these are checks that passed, not checks
demonstrated to fire''---their sensitivity to an actually faulty kernel was never measured,
and a suite that has only ever agreed with good implementations has, by that history alone,
never been tested. Second, the probe checkpoint degraded perplexity by $+157\%$, so the
study explicitly declined to claim power-of-two scales as a deployable mitigation---leaving
the suite's one bitwise-strict check with no servable checkpoint to run on. This paper
measures both.

This paper closes both gaps, and the two answers turn out to be one story
(Fig.~\ref{fig:overview}, bottom). We measure the suite's sensitivity by fault injection:
a reference implementation of the W8A8 pipeline whose accumulator is exact by
construction, nine injected fault families with ground truth known by construction, the
operands replayed from 196 captured layers of Qwen3-1.7B, and every check--fault outcome
predicted in advance: a 63-cell matrix fixed before any data, expanded to 77 cells and
corrected in three by a disclosed two-layer smoke run, then re-pinned before the full run.
The result is a clean scorecard with a damning reading: zero false positives and zero
false negatives against the corrected matrix, yet four of the five epilogue faults are caught by nothing, in any of their
4{,}704 cells. The reason is structural. A single rounding-mode or precision fault moves
an output to a neighbouring representable value---at most one bfloat16 spacing, exactly
one whenever it moves at all---while one spacing is already the smallest tolerance that
admits any legal floating-point difference. Tolerance-based epilogue conformance fails not
because its threshold is badly chosen but because the fault class and the legal class
occupy the same interval. The exit is the check that requires equality instead of a
tolerance, and that check needs power-of-two scales---so deployability of the constraint
stops being an accuracy question and becomes the price of epilogue checkability. We show
the reported $+157\%$ was an artifact of the probe's construction (it rewrote stored
scales without requantizing the weights, a fault of exactly the operand-mismatch family),
and that requantized power-of-two checkpoints carry observed perplexity point estimates
between $-0.28\%$ and $+0.48\%$ at three model sizes while making the two kernels agree
byte for byte.

The measurement discipline matters here, because a sensitivity study can flatter itself
by choosing faults it already catches, and a cost study can flatter itself by comparing
against a strawman. Both measurements therefore run under a pre-registered protocol with
append-only amendments: the fault catalogue, the original 63-cell prediction matrix, all
thresholds, and the analysis rule were fixed before any P5 data; a disclosed two-layer
smoke run then expanded the matrix and corrected three predictions, and the corrected
77-cell matrix was re-pinned before the full execution, with results
reported under both the original and the corrected matrix
(Section~\ref{sec:matrix-corrections}); and the requantization pipeline is validated by
rebuilding the committed baseline checkpoint byte for byte before any constrained arm is
built (Section~\ref{sec:requant}).

Concretely, this paper contributes:
\begin{enumerate}
\item \textbf{A sensitivity measurement for an existing conformance suite}, via nine
injected fault families with constructive ground truth over 8{,}232 layer--fault--regime
cells, scored against a 77-cell prediction matrix (63 cells fixed pre-data; three
corrected after a disclosed smoke run): zero false positives and zero false negatives
against the corrected matrix---75.0\% detection and 55.76\%/11.39\% false-positive
rates under the original---achieved while four of five epilogue faults go entirely
undetected (Section~\ref{sec:sensitivity}).
\item \textbf{A structural explanation and a retraction}: every injected epilogue fault
stays within one bfloat16 spacing of the correct output, so the suite's real-scale
tolerance check is blind to the class by construction; the companion's framing of the
suite as deciding kernel interchangeability is withdrawn and replaced by what the checks
actually establish---preconditions, operand provenance, and one-spacing boundedness.
\item \textbf{Power-of-two scales as a deployable determinism mechanism}: requantized
(not rewritten) power-of-two checkpoints give bitwise per-layer agreement (196/196,
252/252) and byte-identical generation at 1.7B, 8B, and 14B, at measured perplexity point
estimates from $-0.28\%$ to $+0.48\%$ (90\% intervals reaching $+0.71\%$ and
$+0.76\%$)---with the $99.8\%$ decomposition that retires the reported $+157\%$ (Section~\ref{sec:pow2}).
\item \textbf{A byte-exact reconstruction of the quantization pipeline} behind the
studied checkpoints---including the detail that the quantizing division itself happens in
bfloat16---which is what licenses ``the arms differ only in the scale rule'' as an audited
property rather than an intention (Section~\ref{sec:requant}).
\end{enumerate}

The remainder of this paper is organized as follows. Section~\ref{sec:background} states
the pipeline, the alibi, and the related work by treatment axis.
Section~\ref{sec:design} describes the fault-injection design: the reference pipeline,
the fault catalogue, the checks, and the two-stage-pinned prediction matrix.
Section~\ref{sec:sensitivity} reports the sensitivity measurement and its structural
reading. Section~\ref{sec:pow2} reports the power-of-two deployment result: the
requantization gate, determinism at three sizes, the accuracy cost, and the decomposition
of the probe's $+157\%$. Section~\ref{sec:limits} bounds what may be concluded, and
Section~\ref{sec:conclusion} concludes with the strategy this evidence supports.

\section{Background and Related Work}
\label{sec:background}
\label{sec:related}

This section fixes notation and the one theoretical fact everything rests on, then places
the paper among prior work along three axes; Table~\ref{tab:related} summarizes the
comparison.

\subsection{The Pipeline and the Alibi}

A W8A8 linear layer computes $Y = (AW^{\top}) \cdot s_a s_w$, where $A$ is an
$M{\times}K$ int8 activation tile with per-token scales $s_a$, $W$ an $N{\times}K$ int8
weight matrix with per-channel scales $s_w$; the product accumulates in INT32 and the
scaled result rounds to bfloat16. Integer accumulation is exact and order-independent
whenever it cannot overflow, and for the checkpoints studied here the per-product bound
is exactly $16256$: the quantizer divides by $255/2 = 127.5$, which carries the
\emph{weights} to $-128$, while the captured activations stay within $[-127,127]$ at
every one of the 196 layers, so the largest product is $128 \times 127$ and
\begin{equation}
\max_{i,j} |{\rm acc}_{ij}| \;\le\; 16256\,K \;<\; 2^{31}
\quad\text{for all } K \le 132{,}104 .
\label{eq:bound}
\end{equation}
Under (\ref{eq:bound}) any tiling, any split-$K$, any reduction tree produces the
identical INT32 value; the accumulator is above suspicion, and cross-kernel differences
can arise only in the epilogue. The second fact is the power-of-two commutation
\begin{equation}
\mathrm{rnd}(x)\cdot 2^{k} \;=\; \mathrm{rnd}(x \cdot 2^{k})
\label{eq:commute}
\end{equation}
for finite normal values: if all scales are powers of two, the epilogue's two legal
multiplication orderings produce bit-identical float32 and the final cast rounds the same
value once, so bitwise equality becomes \emph{required} rather than expected. Both facts,
their hardware verification, and their failure regimes (subnormals, overflow) are
established in the companion study~\cite{integeralibi2026}; this paper uses them as
given.

\subsection{Related Work by Treatment Axis}

\begin{table}[!t]
\caption{Where this paper sits. ``Bitwise'' = compares outputs at bit granularity;
``ground truth'' = evaluates a test suite against faults whose presence is known by
construction; ``deployable cost'' = measures the accuracy price of a determinism
mechanism on served checkpoints; ``pre-reg.'' = predictions or protocol pinned before
measurement.}
\label{tab:related}
\centering
\footnotesize
\setlength{\tabcolsep}{3pt}
\begin{tabular}{lcccc}
\toprule
 & bitwise & ground truth & deploy.\ cost & pre-reg. \\
\midrule
CRADLE~\cite{pham2019cradle}          &  --      & --  & -- & -- \\
Predoo~\cite{zhang2021predoo}         &  --      & --  & -- & -- \\
Eq.\ checking~\cite{dubey2025volta}   & \fire    & --  & -- & -- \\
SMT tensor cores~\cite{valpey2025smt} & \fire    & --  & -- & -- \\
Dev.\ frameworks~\cite{schlogl2023deviations} & -- & -- & -- & -- \\
Nondet.\ LLM~\cite{yuan2025nondeterminism}    & -- & -- & -- & -- \\
Batch inv.~\cite{he2025batchinvariance}       & \fire & -- & \fire & -- \\
TP-invariant~\cite{tbik2025}          & \fire    & --  & \fire & -- \\
Companion~\cite{integeralibi2026}     & \fire    & --  & -- & \fire \\
\textbf{This paper}                   & \fire    & \fire & \fire & \fire \\
\bottomrule
\end{tabular}
\end{table}

\subsection{Nondeterminism in LLM Inference}
Run-to-run and batch-dependent variance in serving stacks is documented and increasingly
engineered against: numerical deviations across inference
frameworks~\cite{schlogl2023deviations} (2023), nondeterminism in deployed LLM
inference~\cite{yuan2025nondeterminism} (2025), batch-invariant kernels that remove
batch-composition dependence~\cite{he2025batchinvariance} (2025), deterministic inference
across tensor-parallel sizes~\cite{tbik2025} (2026), and verification of served outputs
despite nondeterminism~\cite{karvonen2025difr} (2025). Outside machine learning,
bitwise-reproducible floating-point summation has a longer
history~\cite{collange2015reproducible,ahrens2020reproducible}, using exact
superaccumulator or reproducible-accumulator schemes that neutralize order dependence.
That axis concerns the variance of one implementation
against itself; the axis here is two implementations, each internally deterministic, that
disagree with each other---and the power-of-two condition removes the arithmetic freedom
the epilogue is permitted rather than pinning a reduction order, leaving the reduction to
the alibi.

\subsection{Testing Numerical Kernels}
Differential testing of deep-learning stacks (CRADLE~\cite{pham2019cradle}, 2019) and
precision testing of individual operators (Predoo~\cite{zhang2021predoo}, 2021) search
for inputs that expose disagreement between implementations; equivalence checking can prove or refute equivalence for a defined class of GPU
kernels~\cite{dubey2025volta} (2025), SMT formalizations of tensor-core
semantics~\cite{valpey2025smt} (2025) model the hardware arithmetic and generate inputs
that discriminate among candidate behaviours, and the
rounding behaviour of tensor-core units has been probed
directly~\cite{fasi2021tensorcores} (2021). All of these hold the implementations under
test as the unknown. Our question inverts it: hold the \emph{test suite} as the object
under study, inject faults whose ground truth is known by construction, and measure what
the suite sees---the fault-injection tradition applied to a numerical conformance suite.

\subsection{Constrained Quantization Scales}
Dyadic scales keep pipelines integer-only in HAWQ-V3~\cite{yao2021hawqv3} (2021); RAPQ
fits power-of-two scales for accuracy at low bit-width~\cite{yao2022rapq} (2022);
accumulator-aware quantization guarantees overflow avoidance by
construction~\cite{colbert2023a2q,colbert2025axe} (2023, 2025); and the emulation
literature derives admissible reduction depths from accumulator bit
budgets~\cite{ootomo2024dgemm,abdelfattah2025intmm} (2024, 2025). The same constraint
appears here with a different objective---cross-kernel bitwise agreement---and with its
accuracy cost measured under a pinned protocol rather than optimized.

\subsection{Pre-Registration}
Pre-registration with append-only amendment logs has been proposed for
predictive-modeling research~\cite{hofman2023prereg}, and blind analysis is long-standing
practice in experimental physics~\cite{klein2005blind}. Here it is load-bearing: every
check--fault prediction and every threshold was pinned in writing---the original matrix
before any data, three disclosed corrections re-pinned before the full run---which is
what gives the three post-data corrections of Section~\ref{sec:matrix-corrections} an
audit trail instead of a suspicion.

\section{Fault-Injection Design}
\label{sec:design}

This section builds the measurement instrument: a reference pipeline whose accumulator
is exact by construction (\ref{sec:refpipe}), nine injectable faults
(Table~\ref{tab:faults}), the seven checks under study (\ref{sec:checks}), and the
prediction matrix each cell is scored against
(Table~\ref{tab:matrix}).

\subsection{A Reference Pipeline Exact by Construction}
\label{sec:refpipe}

To know a fault's ground truth, the fault-free pipeline must be exact by construction
rather than by assumption. We compute the accumulator in float64 when the reduction depth
permits---with products bounded by $16256$, a float64 mantissa holds any partial sum
exactly for $K \le 2^{53}/16384$, which covers every layer here---and fall back to INT64
otherwise; the two paths agree bitwise on the real shapes, including matrices whose
weights all sit at $-128$. Operands are not synthetic: we replay the int8 activations
captured from every linear layer of Qwen3-1.7B on the pinned prompts, the same operands
the companion's localization was computed on, at $K$ from 1{,}042 to 12{,}288.

\subsection{Nine Faults}

\begin{table}[!t]
\caption{The fault catalogue. Epilogue faults are single, locally plausible mistakes a
conventionally correct kernel could contain; precondition and operand faults violate
what the exactness argument assumes; the null fault changes nothing, making silence
informative.}
\label{tab:faults}
\centering
\footnotesize
\begin{tabular}{lp{5.3cm}}
\toprule
fault & what it does \\
\midrule
\multicolumn{2}{l}{\emph{epilogue}} \\
F1 scale precision   & scales cast to bfloat16 before multiplying \\
F2 double rounding   & extra rounding to bfloat16 between the two scale multiplies \\
F3 scale order       & $\mathrm{acc}\cdot(s_a s_w)$ in place of $(\mathrm{acc}\cdot s_a)\cdot s_w$ \\
F4 output truncation & output cast truncates toward zero instead of round-to-nearest-even \\
F5 fused order       & per-column scale folded into the reduction, so the sum becomes a float sum \\
\midrule
\multicolumn{2}{l}{\emph{preconditions}} \\
F6 INT32 overflow    & reduction depth inflated until the accumulator wraps \\
F7 above $2^{24}$    & accumulator pushed past exact float32 representability \\
\midrule
F8 null              & nothing \\
F9 operand mismatch  & the two arms receive different int8 operands \\
\bottomrule
\end{tabular}
\end{table}

Table~\ref{tab:faults} lists the catalogue (see also Fig.~\ref{fig:overview}, top). Each
fault runs at three coverage severities---one output element, one percent of elements,
all elements---and under two scale regimes, the checkpoint's own scales and power-of-two
scales. With 196 layers this yields 8{,}232 cells.

\subsection{The Checks, and What Each May Conclude}
\label{sec:checks}

The suite under study has seven checks: (1) the two arms received identical operands;
(2) $16256\,K < 2^{31}$; (3) the maximum accumulator magnitude stays at or below
$2^{24}$; (4) the two accumulators are identical; (5) outputs are bit-identical when all
scales are powers of two; (6) the maximum ULP distance under real scales stays within a
tolerance; (7) a token-level risk check on logit margins. Checks (1)--(5) have
falsifiable outcomes rather than thresholds to tune---(4) admits no legal difference at
all, and (5) turns the rounding argument (\ref{eq:commute}) into a required equality.
Checks (6)--(7) are tolerance-based; we pin the ULP tolerance at one spacing, the
smallest value that admits any legal floating-point difference. Each check's
\emph{applicability} is reported separately from its verdict: a check that cannot apply
to a cell is not a check that passed, and conflating the two is the easiest way to
inflate a suite's apparent coverage.

\subsection{A Prediction Matrix Pinned in Two Stages}
\label{sec:matrix}

The matrix was pinned in two stages, and the distinction is load-bearing. Before any P5
data existed we wrote, for each of 63 check--fault pairs (nine faults by seven checks),
whether the check should fire, stay silent, or does not apply, with the reason: eight
cells predicted to fire, 46 silent, nine not applicable, and five \textsc{predicted
miss}---cells where we \emph{expected} a real defect to pass a check, recorded in
advance precisely so the misses could not later be presented as discoveries. A disclosed
two-layer smoke run then expanded the matrix to the 77 cells of Table~\ref{tab:matrix}
and changed three predictions (Section~\ref{sec:matrix-corrections}): F9 was split by
severity because its one-percent rung perturbs a scale rather than the int8 tensors,
which is what changed two of them, and the split introduced a sixth predicted miss. The
corrected 77-cell matrix was re-pinned before the full 196-layer run; every score in
Section~\ref{sec:sensitivity} is against that corrected matrix, never silently against
the original. Eleven cells are not applicable---all of them check (7), whose sensitivity
a layer-level injection cannot measure at all; we report a suite of six checks measured
and one not, rather than seven.

Completing the matrix surfaced a structural fact that the pre-registration required us to
disclose before measuring, and it is the result of Section~\ref{sec:sensitivity} stated in
advance: every predicted fire came from the precondition and operand faults, and \emph{not
one cell predicted that any check would catch any of the five epilogue faults}. The
amendment records the arithmetic reason---F1 moves a scale by at most $2^{-9}$ relative,
about half a spacing; F2 half a spacing per step; F4 at most one spacing; F3 and F5 were
measured below bfloat16 resolution---and then commits, conditionally and in writing, to the
consequence: if measurement confirmed the expectation, the suite's positioning must be
narrowed from deciding interchangeability to deciding whether the preconditions hold,
whether operands are shared, and whether differences exceed one spacing. The retraction in
Section~\ref{sec:sensitivity} is therefore the discharge of a pre-registered commitment
rather than a concession extracted afterwards, and the tolerance was not chosen to make the
suite look good: it was set from the value the companion had observed. One
bookkeeping note for auditors: the matrix file's prose summary of the predicted misses names
F6 against checks (5) and (6), while the authoritative per-cell records mark F6 against (5)
only, the sixth miss being F9's one-percent rung against (6); the cells are what the suite
was scored against, the summary sentence miscounts, and we report the discrepancy rather
than repairing it silently.

The reading rule was pinned before execution: for each cell, both arms run on the same
captured operands, each check records applicability and verdict separately, and the cell
scores a false negative if a \textsc{should-fire} prediction stays silent, a false positive
if a \textsc{should-not-fire} prediction fires. A cell whose injected fault changed no
output element is excluded from both denominators and reported separately---an unobservable
fault tests nothing, and counting it as a correct silence would credit the suite for a case
it never faced.

\begin{table*}[!t]
\caption{The 77-cell prediction matrix as re-pinned before the full run (the pre-data
original had 63 cells; F9 was one row). \fire{} = should fire, \silent{} = should stay
silent, --- = not applicable; superscript $m$ = \textsc{predicted miss} (five pre-data,
the sixth introduced by the F9 split); superscript $*$ = corrected post-data. Checks: (1) shared operands, (2) INT32 no-overflow, (3) lossless
float32 entry, (4) exact accumulator, (5) power-of-two identity, (6) real-scale
tolerance, (7) token-level risk.}
\label{tab:matrix}
\centering
\footnotesize
\begin{tabular}{lccccccc}
\toprule
fault & (1) & (2) & (3) & (4) & (5) & (6) & (7) \\
\midrule
F1 precision   & \silent & \silent & \silent & \silent & \silent & \silent$^m$ & --- \\
F2 dbl.\ round & \silent & \silent & \silent & \silent & \silent & \silent$^m$ & --- \\
F3 order       & \silent & \silent & \silent & \silent & \silent & \silent     & --- \\
F4 truncate    & \silent & \silent & \silent & \silent & \fire$^{*}$ & \silent$^m$ & --- \\
F5 fused       & \silent & \silent & \silent & \silent & \silent & \silent$^m$ & --- \\
F6 overflow    & \silent & \fire   & \fire   & \fire   & \silent$^m$ & \silent & --- \\
F7 above $2^{24}$ & \silent & \silent & \fire & \silent & \silent & \silent   & --- \\
F8 null        & \silent & \silent & \silent & \silent & \silent & \silent    & --- \\
F9, all elem.  & \fire   & \silent & \silent & \fire   & \fire   & \fire      & --- \\
F9, one elem.  & \fire   & \silent & \silent & \fire   & \fire   & \fire$^{*}$ & --- \\
F9, one pct.   & \fire   & \silent & \silent & \silent$^{*}$ & \silent & \silent$^m$ & --- \\
\bottomrule
\end{tabular}
\end{table*}

\subsection{Three Cells Corrected After Seeing Data}
\label{sec:matrix-corrections}

Three of the 77 cells were changed after a two-layer smoke run contradicted their stated
reasoning, and we report them prominently because the pattern is exactly what post-hoc
fitting produces: all three corrections moved in the suite's favour, together flipping
1{,}278 of the 8{,}232 cells from error to agreement. Under the \emph{original} matrix
the accumulator check shows $75.0\%$ detection (392 false negatives) and the two scale
checks show false-positive rates of $55.76\%$ and $11.39\%$; every headline in
Section~\ref{sec:sensitivity} therefore holds under both matrices. In defence, exactly
two things: each correction repairs an arithmetic error checkable independently of the
outcome (a value exact in float32 need not be exact in bfloat16; rewriting a scale is not
requantizing the weights; cancellation small in absolute terms need not be small
relatively), with the original reasoning preserved verbatim in the artifact
(quoted at the end of this subsection); and the corrected predictions then held with zero error in the full 196-layer
run---including the 194 layers the smoke run never touched, which is out-of-sample
confirmation for them, though not for the two layers the corrections were fitted on. Whether
that suffices is for the reader.

The three cells' original reasoning, preserved in the artifact under
\code{corrected\_post\_data}, reads:

\emph{F4 $\times$ pow2 identity (silent $\to$ fire):}
``Originally predicted silent, on the reasoning that a power-of-two scale makes the
product exactly representable so truncation and round-to-nearest agree. That confused
exact in fp32 with exact in bf16: the accumulator needs seventeen mantissa bits and bf16
carries eight, so the cast still rounds and the two modes still differ. A two-layer smoke
run fired this cell. The correction favours the study, so it is flagged: power-of-two
scales immunise the epilogue against intermediate precision and rounding order---F1 and
F2 are no-ops here---but not against a wrong rounding mode, and check 5 detects that.''

\emph{F9 (one element) $\times$ real-scale tolerance (silent $\to$ fire):}
``Originally predicted silent on a quarter-spacing argument that looked only at the
absolute change. Shifting one int8 activation element moves every accumulator entry in
that row by up to $|w| = 127$, and entries driven near zero by cancellation see that as
an enormous relative change: measured 366 to 384 ulp. Check 6 detects operand differences
far better than the matrix originally allowed.''

\emph{F9 (one percent) $\times$ exact accumulator (fire $\to$ silent):}
``Originally predicted fire, on the reasoning that different operands give different
accumulators. At this rung the difference is in a scale and the int8 tensors are
untouched, so the accumulator is identical and silence is correct. The original
prediction was right for the other two rungs, which is why the cell had to be split.''

\section{How Much Can a Tolerance See?}
\label{sec:sensitivity}

This section reports the sensitivity measurement---first the scorecard, then the
detection table it hides, then the structural reason and what survives of the suite.

\subsection{Result}

Against the corrected matrix, re-pinned before the full run with its three post-smoke
corrections disclosed (Section~\ref{sec:matrix-corrections}), the suite records
\textbf{zero false negatives and zero false positives} over all 8{,}232 cells, and the
null fault F8 never fires in any of its 392 cells. Read alone, a clean bill of health. Table~\ref{tab:sensitivity} shows why it
is not.

\begin{table}[!t]
\caption{Detection by fault. ``Obs.'' counts cells where the injected fault changed at
least one output element; unobservable cells are excluded from the detection denominator
(Section~\ref{sec:matrix}). ``Det.'' counts cells where at least one check fired.}
\label{tab:sensitivity}
\centering
\footnotesize
\setlength{\tabcolsep}{3pt}
\begin{tabular}{lrrrl}
\toprule
fault & cells & obs. & det. & detected by \\
\midrule
\multicolumn{5}{l}{\emph{epilogue}} \\
F1 scale precision      & 1{,}176 & 439 & \textbf{0} & --- \\
F2 double rounding      & 1{,}176 & 448 & \textbf{0} & --- \\
F3 scale order          & 1{,}176 & 205 & \textbf{0} & --- \\
F4 truncation           & 1{,}176 & 972 & 494 & pow2 identity \\
\ \ \emph{real scales}  &   588 & 478 & \textbf{0} & --- \\
\ \ \emph{pow2 scales}  &   588 & 494 & 494 & pow2 identity \\
\midrule
\multicolumn{5}{l}{\emph{preconditions}} \\
F6 overflow             &   392 & 392 & 392 & all three acc.\ checks \\
F7 above $2^{24}$       &   392 & 392 & 392 & fp32 entry \\
\midrule
F8 null                 &   392 &   0 &   0 & --- \\
F9 operand mismatch     & 1{,}176 & 984 & 1{,}176 & operands $+$ 3 others \\
\bottomrule
\end{tabular}
\end{table}

Four of the five epilogue faults---4{,}704 cells---are detected by \emph{no} check, in
either scale regime, at any severity, including the severity that corrupts every output
element. The fifth (truncation) is detected in 494 cells, all under power-of-two scales
and all by the power-of-two identity check; under the checkpoint's own scales the same
fault is observable in 478 cells and detected in none. The preconditions behave in the
opposite way: both violations are caught in every cell, and the operand fault in all
1{,}176---including the 192 cells where it changed no output, because comparing operands
does not depend on the fault becoming visible downstream. The contrast is the useful
result: the suite is reliable about its own assumptions and about operand provenance, and
uninformative about the epilogue.

\subsection{Why: One Spacing Is the Whole Budget}

\begin{figure}[!t]
\centering
\includegraphics[width=\linewidth]{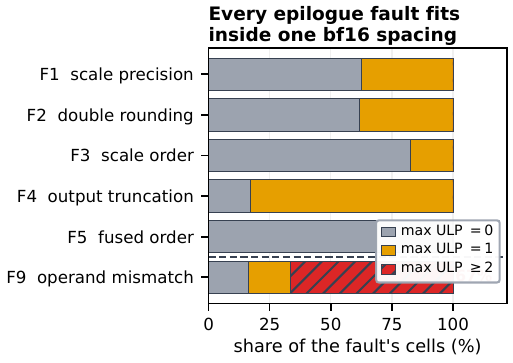}
\caption{Where each fault lands, as the real-scale tolerance check sees it. Above the dashed
rule are the five epilogue faults: every one produces a maximum ULP distance of exactly $0$
or $1$---never $2$---across all 5{,}880 cells, so the entire class sits inside the
one-spacing interval that legal implementations are allowed to occupy. Below it, the operand
fault F9 reaches distances from 68 to 35{,}571 in $67\%$ of its cells. A tolerance
partitions the fault space at the wrong boundary.}
\label{fig:ulp}
\end{figure}

The reason is structural, not an unlucky threshold (Fig.~\ref{fig:ulp}). Each epilogue
fault changes how a single value is rounded or in what order two scalars multiply; either
way the result is a neighbouring representable bfloat16 value, so the output moves by one
step and not by two. The measurement is unambiguous: across all 5{,}880 epilogue cells
the maximum ULP distance takes only the values $0$ and $1$, and in every cell where the
fault is observable at all it is exactly $1$. The operand fault reaches 68 to 35{,}571 in
the same metric. Tightening the tolerance is not available---one spacing is already the
smallest tolerance admitting any legal difference, and a tolerance of zero \emph{is} the
power-of-two check, which requires a different checkpoint rather than a different
threshold. This also explains the one detection: under power-of-two scales,
(\ref{eq:commute}) makes agreement a required equality, and truncation survives that
change---an accumulator needing seventeen mantissa bits still rounds into bfloat16's
eight, so truncating and rounding-to-nearest still differ---which is why F4 is the one
epilogue fault the suite sees, and only there.

\subsection{What the Suite Establishes; a Retraction}

The companion's framing of these checks as a procedure for deciding kernel
\emph{interchangeability} does not survive this measurement, and we retract it---as
amendment A-10.4 committed us to do, in writing and before the suite ran, should the
measurement come out this way (Section~\ref{sec:matrix}). What the
suite establishes is narrower and worth stating exactly: that the preconditions of the
exactness argument hold on the evaluated inputs; that the two implementations received
the same operands, so a passing comparison is not an artifact of provenance; and that
observed differences stay within one output spacing. A kernel containing any of F1, F2,
F3, or F5 passes every check in the suite. Under the original (uncorrected) matrix the
same data read as $75.0\%$ detection for the accumulator check and $55.76\%$ /
$11.39\%$ false-positive rates for the two scale checks; the conclusion is unchanged
either way, since the corrections concern which cells were \emph{predicted} to fire, not
which checks fired. The token-level check remains untested: a layer-level injection
supplies no logit margins, so its eleven cells are not applicable and its sensitivity is
unmeasured.

\section{Power-of-Two Scales as a Deployable Condition}
\label{sec:pow2}

Section~\ref{sec:sensitivity} leaves one check that can see inside the epilogue, and it
requires power-of-two scales---so the constraint's deployability is the price of
epilogue checkability. This section shows the reported $+157\%$ was the probe's
construction, not the constraint (\ref{sec:requant}), then measures determinism
(\ref{sec:bitwise}) and cost (\ref{sec:cost}) on requantized checkpoints at three sizes.

\subsection{Requantizing Rather than Rewriting}
\label{sec:requant}

The companion's probe rewrote the stored \code{weight\_scale} tensors and left the int8
weights untouched, so the weights remained quantized for the old scales and were
interpreted through new ones---a weight--scale mismatch, which is a fault of exactly the
F9 family. The correct construction requantizes: from the parent bfloat16 weights, each
channel's scale is computed by the checkpoint's own rule and then constrained, and the
int8 weights are recomputed under the constrained scale. The step that licenses the
comparison is the gate: the reconstruction must reproduce the committed baseline byte for
byte under the \emph{unconstrained} rule before any constrained arm is built. Reverse
engineering found the pipeline divides by $255/2=127.5$, stores scales in bfloat16, and
performs the quantizing division itself in bfloat16---eight mantissa bits, moving
quotients by up to $0.749$ with sign varying inside one channel. With that convention the
rebuilt checkpoint hashes identically to the original across all 196 layers; a first
attempt that divided in float32 left $6.42\%$ of the int8 weights different and was
rejected by exactly this gate. The power-of-two arms are the same pipeline with the scale
snapped to $2^{\mathrm{round}(\log_2 s)}$ (nearest) or $2^{\lceil \log_2 s \rceil}$
(ceiling); powers of two are exactly representable in bfloat16, so no further rounding
intervenes.

\subsection{Determinism at Three Sizes}
\label{sec:bitwise}

\begin{table}[!t]
\caption{Cross-kernel agreement at two granularities. Per-layer: identical captured
operands through both kernels, over all linear layers. End to end: greedy generation on 8
pinned prompts of 64 tokens, ``identical'' meaning the token sequence matches byte for byte.
At 14B the capture path dequantizes to bfloat16 and exceeds the 24\,GB card, so no
per-layer figure exists.}
\label{tab:e2e}
\centering
\footnotesize
\begin{tabular}{lcccc}
\toprule
 & \multicolumn{2}{c}{per-layer} & \multicolumn{2}{c}{end to end} \\
\cmidrule(lr){2-3}\cmidrule(lr){4-5}
model & own & pow2 & own & pow2 \\
\midrule
Qwen3-1.7B & 8/196  & \textbf{196/196} & 0/8 & \textbf{8/8} \\
Qwen3-8B   & 10/252 & \textbf{252/252} & 0/8 & \textbf{8/8} \\
Qwen3-14B  & ---    & ---              & 0/8 & \textbf{8/8} \\
\bottomrule
\end{tabular}
\end{table}

Under the requantized nearest arm the two kernels agree bitwise at every linear layer and end to end: byte-identical token sequences on all eight
pinned prompts at all three model sizes, against zero of eight under the checkpoints' own
scales (Table~\ref{tab:e2e}). The kernel choice stops being observable in the output.

\subsection{The Cost, and the Decomposition of $+157\%$}
\label{sec:cost}

\begin{figure}[!t]
\centering
\includegraphics[width=\linewidth]{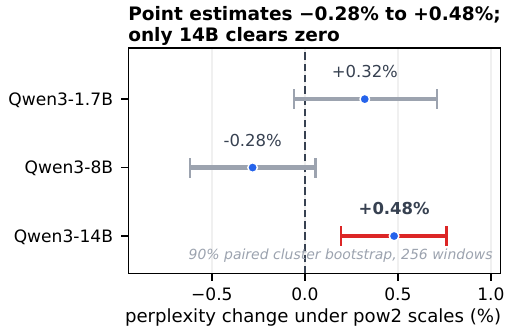}
\caption{Accuracy cost of the nearest-power-of-two constraint: relative perplexity change
with $90\%$ paired cluster-bootstrap intervals over 256 pinned WikiText windows, 10{,}000
draws. The interval covers zero at 1.7B and 8B and excludes it at 14B (drawn in red); point
estimates span $-0.28\%$ to $+0.48\%$, and the interval upper ends reach $+0.71\%$
(1.7B) and $+0.76\%$ (14B). Throughput at 1.7B moves between $-9.6\%$ and $+1.4\%$
(prefill) and $-1.6\%$ and $+0.9\%$ (decode) across six batch--length settings on a shared
machine; 8B and 14B throughput was not measured.}
\label{fig:cost}
\end{figure}

Fig.~\ref{fig:cost} gives the cost. We state it as observed point estimates---$-0.28\%$
to $+0.48\%$, with $90\%$ intervals extending to $+0.71\%$ at 1.7B and $+0.76\%$ at
14B---and neither as ``indistinguishable from zero,'' which is false at 14B, where the
interval excludes zero, nor as a bound, which the interval upper ends do not support. The three sizes ran under
one pinned protocol with the outcome reported regardless of sign; the negative 8B point
estimate is consistent with noise, not evidence of improvement.

The decomposition settles the $+157\%$. The requantized nearest arm costs $+0.32\%$,
so the probe's weight--scale mismatch accounts for $99.8\%$ of its degradation. Within
the remainder, the ceiling arm---which cannot clip, since it never shrinks a
scale---costs \emph{more} than nearest ($+0.54\%$): clipping a few outliers in exchange
for finer resolution is mildly beneficial, opposite to the pre-registered hypothesis. The
pre-registered gate for a fourth search arm (run only if the ceiling cost exceeded
$5\%$) resolved to not-run.

What the constraint buys is stated in Fig.~\ref{fig:overview}, bottom right: cross-kernel bitwise
determinism as a serving configuration, and---the reason it belongs in this paper---the
condition under which this suite's epilogue comparison stops being a tolerance: under the
stated arithmetic preconditions, power-of-two scales replace the tested one-spacing check
with a required equality. The constraint converts the comparison this suite cannot win
into one it can.

\section{Limitations}
\label{sec:limits}

\textbf{The faults are ours.} Zero false negatives is a statement about nine chosen fault
families; a real defect of a form we did not anticipate---especially one moving an output
by more than one spacing through a mechanism other than operand corruption---is outside
the measurement. The blindness result needs no such caveat: it is an existence proof, and
four concrete fault families establish it.
\textbf{One model family, one kernel pair.} All evidence concerns Qwen3 at three sizes
and the CUTLASS--Triton pair inside vLLM; a pre-registered search for a second kernel
pair meeting the isolation criteria examined ten serving engines and found none, recorded
as a negative result in the amendment log. Above 14B, tensor parallelism introduces
cross-device reductions that confound the single-kernel treatment.
\textbf{14B evidence is end-to-end only}, the capture path exceeding the 24\,GB card; we
did not work around it with a second GPU, which would have changed capture conditions
relative to the other sizes.
\textbf{Throughput was measured only at 1.7B}, on a shared machine; the ceiling arm was
built only at 1.7B.
\textbf{The token-level check is unmeasured}, and \textbf{three matrix cells were
corrected post-data}, all in the suite's favour (both denominators reported,
Section~\ref{sec:matrix-corrections}).
\textbf{No decode-regime per-layer capture} exists; the end-to-end sequences span
prefill and 64 decode steps, so decode is covered by the byte-identity result only.

\section{Conclusion and Future Work}
\label{sec:conclusion}

A one-spacing tolerance cannot certify bitwise epilogue equivalence: every fault that
stays inside one output spacing---where single rounding-mode and precision faults live by
construction---passes it, and this holds at the smallest tolerance that admits any legal
difference, so the four fault families of this study go unseen. The suite is not thereby
useless; it is misdescribed. It
reliably certifies preconditions, operand provenance, and one-spacing boundedness, and
we retract the stronger interchangeability framing. The constructive half is that the
one check requiring equality instead of tolerance is now deployable: requantized
power-of-two scales make two production kernels agree byte for byte at three model
sizes, at observed perplexity costs between $-0.28\%$ and $+0.48\%$ (interval upper ends
$+0.71\%$ and $+0.76\%$). The strategy this
evidence supports: a practitioner who wants a bitwise-equality contract for this epilogue,
under the stated arithmetic preconditions, should not tighten a tolerance; they should
remove the scale-application freedom the tolerance was accommodating, which is what the
power-of-two constraint does.

Future work follows the recorded gaps: a second kernel pair (the blocking isolation
criteria are documented; ROCm's aiter path is the cleanest candidate found), per-layer
capture at 14B and in the decode regime, sensitivity of the token-level check under an
end-to-end injection, and fault families beyond the nine---in particular any mechanism
that moves outputs by more than one spacing without touching operands, which would test
the boundary this paper draws.

\section*{Artifact Statement}

The pre-registration amendments governing this study (A-10 through A-12, append-only), with
an English summary of exactly the amendments this paper relies on alongside the
authoritative record, the
77-cell prediction matrix with the three corrected cells' original reasoning preserved
verbatim under \code{corrected\_post\_data} flags, the fault injector and conformance
checks with their regression tests, the requantization pipeline with its byte-exact
reproduction gate, all measurement artifacts with SHA-256 digests, and the per-run records
including the first-run and disk-exhaustion failure logs are maintained in the same
version-controlled repository as the companion study's artifacts, with a linear commit
history and separate plan-and-code pins and result commits. The repository is private at the
time of writing and will be made public with the announcement of this paper, at which point
its URL and the manuscript and result commit hashes will be listed here. Known gaps are
enumerated in Section~\ref{sec:limits}.

\bibliographystyle{IEEEtran}
\bibliography{refs}

\end{document}